\documentclass[11pt,a4paper]{article}
\usepackage[nohyperref]{emnlp-ijcnlp-2019}
\usepackage{times}
\usepackage{latexsym}
\usepackage{linguex}
\usepackage{enumitem}
\usepackage{booktabs}
\usepackage{graphicx}

\newcommand{\mc}[3]{\multicolumn{#1}{#2}{#3}}

\usepackage{url}

\aclfinalcopy 

\title{Analysing Coreference in Transformer Outputs}

\author{~~~~~Ekaterina Lapshinova-Koltunski \hspace{0.4cm} Cristina Espa{\~{n}}a-Bonet  \hspace{0.4cm} Josef van Genabith \\
  \hspace{1.8cm}Saarland University \hspace{2.2cm} Saarland University  \hspace{0.8cm} Saarland University \\
   \hspace{7.2cm}  DFKI GmbH  \hspace{2.2cm} DFKI GmbH  \\
  \texttt{e.lapshinova@mx.uni-saarland.de} \\
   \texttt{\{cristinae,Josef.Van\_Genabith\}@dfki.de} \\}

\date{}

\begin{document}
\maketitle
\begin{abstract}
We analyse coreference phenomena in three neural machine translation systems trained with different data settings with or without access to explicit intra- and cross-sentential anaphoric information. 
We compare system performance on two different genres: news and TED talks.
To do this, we manually annotate (the possibly incorrect) coreference chains in the MT outputs and evaluate the coreference chain translations. We define an error typology that aims to go further than pronoun translation adequacy and includes types such as incorrect word selection or missing words.  
The features of coreference chains in automatic translations are also compared to those of the source texts and human translations. The analysis shows stronger potential translationese effects in machine translated outputs than in human translations. 
\end{abstract}

\section{Introduction}\label{sec:intro}

In the present paper, we analyse coreference in the output of three neural machine translation systems (NMT) that were trained under different settings. We use a transformer architecture~\cite{vaswani2017attention} and train it on corpora of different sizes with and without the specific coreference information. Transformers are the current state-of-the-art in NMT~\citep{barraultEtAl:2019} and are solely based on attention, therefore, the kind of errors they produce might be different from other architectures such as CNN or RNN-based ones.  Here we focus on one architecture to study the different errors produced only under different data configurations.

Coreference is an important component of discourse coherence which is achieved in how discourse entities (and events) are introduced and discussed. Coreference chains contain mentions of one and the same discourse element throughout a text. These mentions are realised by a variety of linguistic devices such as pronouns, nominal phrases (NPs) and other linguistic means. As languages differ in the range of such linguistic means~\citep{LapshinovaEtAl2019,KunzLapshinova2015,NovakNedoluzhko2015,KunzSteiner2012} and in their contextual restrictions~\citep{SleGecco}, these differences give rise to problems that may result in incoherent (automatic) translations. We focus on coreference chains in English-German translations belonging to two different genres. In German, pronouns, articles and adjectives (and some nouns) are subject to grammatical gender agreement, whereas in English, only person pronouns carry gender marking. An incorrect translation of a pronoun or a nominal phrase may lead to an incorrect relation in a discourse and will destroy a coreference chain. 

Recent studies in automatic coreference translation have shown that dedicated systems can lead to improvements in pronoun translation~\cite{clPronoun:2016,discoMT:2017}. However, standard NMT systems work at sentence level, so improvements in NMT translate into improvements on pronouns with intra-sentential antecedents, but the phenomenon of coreference is not limited to anaphoric pronouns, and even less to a subset of them. Document-level machine translation (MT) systems are needed to deal with coreference as a whole. Although some attempts to include extra-sentential information exist \cite{wangEtal:2017,voitaEtal:2018,jeanCho:2019,junczys:2019}, the problem is far from being solved. Besides that, some further problems of NMT that do not seem to be related to coreference at first glance (such as translation of unknown words and proper names or the hallucination of additional words) cause coreference-related errors. 


In our work, we focus on the analysis of complete coreference chains, manually annotating them in the three translation variants. We also evaluate them from the point of view of coreference chain translation. The goal of this paper is two-fold. On the one hand, we are interested in various properties of coreference chains in these translations. They include total number of chains, average chain length, the size of the longest chain and the total number of annotated mentions. These features are compared to those of the underlying source texts and also the corresponding human translation reference. On the other hand, we are also interested in the quality of coreference translations. Therefore, we define a typology of errors, and and chain members in MT output are annotated as to whether or not they are correct. The main focus is on such errors as gender, number and case of the mentions, but we also consider wrong word selection or missing words in a chain. Unlike previous work, we do not restrict ourselves to pronouns. Our analyses show that there are further errors that are not directly related to coreference but consequently have an influence on the correctness of coreference chains. 

The remainder of the paper is organised as follows. Section \ref{sec:relatedwork} introduces the main concepts and presents an overview of related MT studies. Section~\ref{sec:experiments} provides details on the data, systems used and annotation procedures. Section~\ref{sec:analyses} analyses the performance of our transformer systems on coreferent mentions.
Finally we summarise and draw conclusions in Section~\ref{sec:conclusions}.

\section{Background and Related Work}\label{sec:relatedwork}
\subsection{Coreference}\label{sec:coref}
Coreference is related to cohesion and coherence. The latter is the logical flow of inter-related ideas in a text, whereas cohesion refers to the text-internal relationship of linguistic elements that are
overtly connected via lexico-grammatical devices across sentences~\citep{HallidayHasan1976}. As stated by \citet[p. 3]{Hardmeier2012}, this connectedness of texts 
implies dependencies between sentences. And if these dependencies are neglected in translation, 
the output text no longer has the property of connectedness which makes a sequence of sentences a text. 
Coreference expresses identity to a referent mentioned in another textual part (not necessarily in neighbouring sentences) contributing to text connectedness. 
An addressee is following the mentioned referents and identifies them when they are repeated. Identification of certain referents 
depends not only on a lexical form, but also on other linguistic means, e.g. articles or modifying pronouns~\citep{Kibrik2011}. The use of these 
is influenced by various factors which can be language-dependent (range of linguistic means available in grammar) 
and also context-independent (
pragmatic situation, genre). Thus, 
the means of expressing reference differ across languages and genres. This has been shown by some studies in the area of contrastive linguistics~\citep{SleGecco,KunzLapshinova2015,KunzSteiner2012}. 
Analyses in cross-lingual coreference resolution~\citep{Grishina2017,GrishinaStede2015,NovakZabokrtsky2014,GreenEtAL2011} show that there are still unsolved problems that should be addressed.
	
\subsection{Translation studies}\label{sec:translationstudies}
Differences between languages and genres in the 
linguistic means expressing reference are 
important for translation, as the choice of an appropriate referring expression in the target language 
poses challenges for both human and machine translation. In translation studies, there is a number of corpus-based works 
analysing these differences in translation. 
However, most of them are restricted to individual phenomena within coreference. For instance,~\citet{ZinsmeisterEtAl2012} analyse abstract anaphors in English-German translations. To our knowledge, they do not consider chains. \citet{LapshinovaHardmeier2017} in their contrastive analysis of potential coreference chain members in English-German translations, describe transformation patterns that contain different types of referring expressions. However, the authors rely on automatic tagging and parsing procedures and do not include chains into their analysis. The data used by \citet{NovakNedoluzhko2015} and~\citet{Novak2018} contain manual chain annotations. The authors focus on different categories of anaphoric pronouns 
in English-Czech translations, though not paying attention to chain features (e.g. their number or size).

Chain features are considered in a contrastive analysis by~\citet{SleGecco}. Their study concerns different 
phenomena in a variety of genres in English and German comparable texts. Using contrastive interpretations, they suggest preferred translation strategies from English into German, i.e. translators should use demonstrative pronouns instead of personal pronouns (e.g. {\sl dies/das}
instead of {\sl es/it}) when translating from English into German and vice versa. However, corpus-based 
studies show that translators do not necessarily apply such strategies. Instead, they often preserve the source language anaphor's categories~\citep[as shown e.g. by][]{ZinsmeisterEtAl2012} which results in the shining through effects~\citep{Teich2003}. Moreover, due to the tendency of translators to explicitly realise meanings in translations that were implicit in the source texts~\citep[explicitation effects,][]{BlumKulka1986}, translations are believed to contain more (explicit) referring expressions, and subsequently, more (and longer) coreference chains. 

Therefore, in our analysis, we focus on the chain features related to the  phenomena of shining through and explicitation. These features include number of mentions, number of chains, average chain length and the longest chain size. Machine-translated texts are compared to their sources and the corresponding human translations in terms of these features. We expect to find shining through and explicitation effects in automatic translations. 

\subsection{Coreference in MT}\label{sec:corefinmt}

As explained in the introduction, several recent works tackle the automatic translation of pronouns and also coreference~\citep[for instance,][]{VoigtJurafsky2012, miculicichEtAl:2017} and this has, in part, motivated the creation of devoted shared tasks and test sets to evaluate the quality of pronoun translation \cite{clPronoun:2016,DiscoMT2017,guillouEtAl:2018,bawdenEtal:2018}.

But coreference is a wider phenomenon that affects more linguistic elements. Noun phrases also appear in coreference chains but they are usually studied under coherence and consistency in MT. \citet{xiongEtAl:2015} use topic modelling to extract coherence chains in the source, predict them in the target and then promote them as translations. 
\citet{martinezEtal:2017} use word embeddings to enforce consistency within documents. Before these works, several methods to post-process the translations and even including a second decoding pass were used \cite{carpuat:2009,xiaoEtAl:2011,tureEtAl:2012,martinezEtAt:2014}.

Recent NMT systems that include context deal with both phenomena, coreference and coherence, but usually context is limited to the previous sentence, so chains as a whole are never considered.
\citet{voitaEtal:2018} encode both a source and a context sentence and then combine them to obtain a context-aware input. The same idea was implemented before by \citet{tiedemannScherrer:2017} where they concatenate a source sentence with the previous one to include context. Caches \cite{tuEtal:2018}, memory networks  \cite{marufHaffari:2018} and hierarchical attention methods \cite{miculicichEtal:2018} allow to use a wider context.
Finally, our work is also related to \citet{stojanovskiFraser:2018} and  \citet{stojanovskiFraser:2019} where their oracle translations are similar to the data-based approach we introduce in Section~\ref{sec:nmt}.



\section{Systems, Methods and Resources}
\label{sec:experiments}

\subsection{State-of-the-art NMT}
\label{sec:nmt}

Our NMT systems are based on a transformer architecture \cite{vaswani2017attention} as implemented in the {\tt Marian} toolkit~\cite{mariannmt} using the \emph{transformer big} configuration.

\addtolength{\tabcolsep}{-0.2em} 
\begin{table}
\begin{tabular}{lrrr}
\midrule
 & \# lines~~~ & \mc{1}{c}{S1, S3} & \mc{1}{c}{S2} \\
\midrule
Common Crawl     & 2,394,878 & x1 & x4 \\ 
Europarl        & 1,775,445 & x1 & x4 \\ 
News Commentary &   328,059 & x4 & x16\\ 
Rapid           & 1,105,651 & x1 & x4 \\ 
ParaCrawl Filtered & 12,424,790 & x0 & x1\\ 
\midrule
\end{tabular}
\caption{Number of lines of the corpora used for training the NMT systems under study. The 2nd and 3rd columns show the amount of oversampling used.}
\label{tab:corpusEnDe}
\end{table} 
\addtolength{\tabcolsep}{1pt} 

We train three systems (S1, S2 and S3) with the corpora summarised in Table~\ref{tab:corpusEnDe}%
.\footnote{All corpora are freely available for the WMT news translation task and can be downloaded from \url{http://www.statmt.org/wmt19/translation-task.html}} The first two systems are transformer models trained on different amounts of data (6M vs. 18M parallel sentences as seen in the Table). The third system includes a modification to consider the information of full coreference chains throughout a document augmenting the sentence to be translated with this information and it is trained with the same amount of sentence pairs as S1. A variant of the S3 system participated in the news machine translation of the shared task held at WMT 2019~\cite{espanaEtAl:WMT:2019}.

\paragraph{S1} is trained with the concatenation of Common Crawl, Europarl, a cleaned version of Rapid and the News Commentary corpus. We oversample the latter in order to have a significant representation of data close to the news genre in the final corpus.

\paragraph{S2} uses the same data as S1 with the addition of a filtered portion of Paracrawl. This corpus is known to be noisy, so we use it to create a larger training corpus but it is diluted by a factor 4 to give more importance to high quality translations.

\paragraph{S3} S3 uses the same data as S1, but this time enriched with the cross- and intra-sentential coreference chain markup as described below
.\footnote{Paracrawl has document boundaries but with a mean of 1.06 sent/doc which makes it useless within our approach.} The information is included as follows.


Source documents are annotated with coreference chains using the neural annotator of {\tt Stanford CoreNLP}~\cite{standfordCoreNLP}%
\footnote{This system achieves a precision of $80\%$ and recall of $70\%$ on the CoNLL 2012 English Test Data \citep{clark2016deep}. \citet{voitaEtal:2018} estimated an accuracy of $79\%$ on the translation of the pronoun \emph{it}.}. The tool detects pronouns, nominal phrases and proper names as mentions in a chain. For every mention, {\tt CoreNLP} extracts its gender (male, female, neutral, unknown), number (singular, plural, unknown), and animacy (animate, inanimate, unknown). This information is not added directly but used to enrich the single sentence-based MT training data by applying a set of heuristics 
implemented in {\tt DocTrans}%
\footnote{\url{https://github.com/cristinae/DocTrans/}}:

%
%

\begin{enumerate}
\itemsep-0.2em 
 \item We enrich \emph{pronominal mentions} with the exception of "I" with the head (main noun phrase) of the chain. The head is cleaned by removing articles and Saxon genitives and we only consider heads with less than 4 tokens in order to avoid enriching a word with a full sentence

 \item We enrich \emph{nominal mentions} including \emph{proper names} with the gender of the head
 \item The head itself is enriched with she/he/it/they depending on its gender and animacy
\end{enumerate}



The enrichment is done with the addition of tags as shown in the examples:
\begin{itemize}
 \item I never cook with \emph{$<$b\_crf$>$ salt $<$e\_crf$>$} it.
 \vspace{-0.8em}
 \item \emph{$<$b\_crf$>$ she $<$e\_crf$>$} Biles arrived late. 
\end{itemize}

\noindent
In the first case heuristic 1 is used, \emph{salt} is the head of the chain and it is prepended to the pronoun. The second example shows a sentence where heuristic 2 has been used and the proper name \emph{Biles} has now information about the gender of the person it is referring to.

Afterwards, the NMT system is trained at sentence level in the usual way.
The data used for the three systems is cleaned, tokenised, truecased with Moses scripts
\footnote{\url{https://github.com/moses-smt/mosesdecoder/tree/master/scripts}} 
and BPEd with subword-nmt%
\footnote{\url{https://github.com/rsennrich/subword-nmt}}  
using separated vocabularies with 50\,k subword units each. The validation set ($news2014$) and the test sets described in the following section are pre-processed in the same way.

\begin{table*}
\centering
\begin{tabular}{l rrrrr c rrrrr}
\toprule
    & \mc{5}{c}{\textbf{news}} & & \mc{5}{c}{\textbf{TED}} \\
     \cmidrule(lr){2-6}        \cmidrule(lr){8-12} 
 & tokens &   \#ment. & \#chains  & avg.   & max. &  & tokens &   \#
 ment. & \#chains  & avg. & max. \\
 &        &         &         & length     & length&  &        &   &         & length  & length  \\
\midrule
\bf src              & 9,862 & 782 & 176 & 5.1 & 15.8 & & 11,155 & 1,042 & 338 &  2.9 & 34.7 \\ 
\bf src$_{\rm CoreNLP}$ & 10,502 & 915 & 385 & 2.3 & 13.2 & & 11,753 &  989 & 407 &  2.4 & 30.3\\  
\bf ref              & 9,728  & 851 & 233 & 3.8 & 14.5 & &  10,140 & 916 & 318 &  2.8 & 38.0 \\  
\midrule
\bf S1            & 9,613 & 1,216 &302 & 4.2 & 17.2 & & 10,547 & 1,270 & 293 & 4.5 & 47.0 \\  
\bf S2            & 9,609 & 1,218 &302 & 4.4 & 17.3 & & 10,599 & 1,268 & 283 & 4.6 & 51.7 \\  
\bf S3            & 9,589 & 1,174 &290 & 4.3 & 16.2 & & 10,305 & 1,277 & 280 & 4.7 & 47.0 \\  
\bottomrule
\end{tabular}
\caption{Statistics on coreference features for news and TED texts considered.  }
\label{tab:statsText}
\end{table*}

\subsection{Test data under analysis}\label{sec:data}

As one of our aims is to compare coreference chain properties in automatic translation with those of the source texts and human reference, we derive data from ParCorFull, an English-German corpus annotated with full coreference chains~\citep{LapshinovaEtAl2018LREC}.\footnote{Available at \url{https://lindat.mff.cuni.cz/repository/xmlui/handle/11372/LRT-2614}} The corpus contains ca. 160.7 thousand tokens manually annotated with about 14.9 thousand mentions and 4.7 thousand coreference chains. For our analysis, we select a portion of English news texts and TED talks from ParCorFull and translate them with the three NMT systems described in~\ref{sec:nmt} above. As texts considerably differ in their length, we select 17 news texts (494 sentences) and four TED talks (518 sentences).
The size (in tokens) of the total data set under analysis -- source (src) and human translations (ref) from ParCorFull and the automatic translations produced within this study (S1, S2 and S3) are presented in Table~\ref{tab:statsText}.


Notably, automatic translations of TED talks contain more words than the corresponding reference translation, which means that machine-translated texts of this type have also more potential tokens to enter in a coreference relation, and potentially indicating a shining through effect. The same does not happen with the news test set. 

\subsection{Manual annotation process}\label{sec:anno}

The English sources and their corresponding human translations into German were already manually annotated for coreference chains. We follow the same scheme as \citet{LapshinovaHardmeier2017guide} to annotate the MT outputs with coreference chains. This scheme allows the annotator to define each markable as a certain mention type (pronoun, NP, VP or clause). The mentions can  be defined further in terms of their cohesive function (antecedent, anaphoric, cataphoric, comparative, substitution, ellipsis, apposition). Antecedents can either be marked as simple or split or as entity or event. The annotation scheme also includes pronoun type (personal, possessive, demonstrative, reflexive, relative) and
modifier types of NPs (possessive, demonstrative, definite article, or none for proper names), 
see~\cite{LapshinovaEtAl2018LREC}
for details. The mentions referring to the same discourse item are linked between each other. We use the annotation tool MMAX2~\citep{MullerStrube2006} which was also used for the  annotation of ParCorFull. 

In the next step, chain members are annotated for their correctness. For the incorrect translations of mentions, we include the following error categories: \emph{gender}, \emph{number}, \emph{case}, \emph{ambiguous} and \emph{other}. The latter category is open, which means that the annotators can add their own error types during the annotation process. With this, the final typology of errors also considered \emph{wrong named entity}, \emph{wrong word}, \emph{missing word}, \emph{wrong syntactic structure}, \emph{spelling error} and \emph{addressee reference}.

The annotation of machine-translated texts was integrated into a university course on discourse phenomena. Our annotators, well-trained students of linguistics, worked in small groups on the assigned annotation tasks (4-5 texts, i.e. 12-15 translations per group). At the beginning of the annotation process, the categories under analysis were discussed within the small groups and also in the class. The final versions of the annotation were then corrected by the instructor. 


\section{Results and Analyses}\label{sec:analyses}

\subsection{Chain features}\label{sec:chainfeat}

First, we compare the distribution of several chain features in the three MT outputs, their source texts and the corresponding human translations.

Table~\ref{tab:statsText} shows that, overall, all machine translations contain a greater number of annotated mentions in both news texts and TED talks than in the annotated source (\textit{src} and \textit{src$_{\rm CoreNLP}$}) and reference (\textit{ref}) texts. Notice that src$_{\rm CoreNLP}$ ---where coreferences are not manually but automatically annotated with {\tt CoreNLP}--- counts also the tokens that the mentions add to the sentences, but not the tags. The larger number of mentions may indicate a strong explicitation effect observed in machine-translated texts. 
Interestingly, {\tt CoreNLP} detects a similar number of mentions in both genres, while human annotators clearly marked more chains for TED than for news. Both genres are in fact quite different in nature; whereas only $37\%$ of the mentions are pronominal in news texts (343 out of 915), the number grows to $58\%$ for TED (577 out of 989), and this could be an indicator of the difficulty of the genres for NMT systems. 

There is also a variation in terms of chain number between translations of TED talks and news. While automatic translations of news texts contain more chains than the corresponding human annotated sources and references, machine-translated TED talks contain less chains than the sources and human translations. However, there is not much variation between the chain features of the three MT outputs. The chains are also longer in machine-translated output than in reference translations as can be seen by the number of mentions per chain and the length of the longest chain.


\begin{table*}
\begin{tabular}{l ccccc c ccccc}
\toprule
    & \mc{2}{c}{\textbf{news}$_{all}$} & \mc{2}{c}{\textbf{news}$_{coref}$} & & & \mc{2}{c}{\textbf{TED}$_{all}$} & \mc{2}{c}{\textbf{TED}$_{coref}$} \\
\cmidrule(lr){2-3} \cmidrule(lr){4-5}\cmidrule(lr){8-9}\cmidrule(lr){10-11}
&   BLEU   & MTR  &    BLEU & MTR & \#mention err.  & & BLEU   & MTR  &    BLEU &  MTR  & \#mention err.  \\ 
\midrule
\bf S1~~          & 30.68 &  55.87 & 30.07 & 55.84 & 117 (9.6\%)  & & 31.99 & 57.91 & 31.70& 58.06& 84  (6.6\%) \\  
\bf S2            & 31.47 &  56.88 & 30.83 & 56.68 &  86 (7.1\%)  & & 32.36 & 58.22 & 32.81& 59.73& 105 (8.3\%)\\  
\bf S3            & 30.35 &  55.26 & 29.89 & 55.24 & 121 (10.3\%) & & 32.67 & 58.84 & 32.84& 58.85 & 83 (6.5\%)\\  
\bottomrule
\end{tabular}
\caption{BLEU and METEOR (MTR) scores for the 3 systems on our full test set ($all$) and the subset of sentences where coreference occurrs ($coref$). The number of erroneous mentions is shown for comparison.}
\label{tab:evalMT}
\end{table*} 

\subsection{MT quality at system level}
\label{sec:evalMT}

We evaluate the quality of the three transformer engines with two automatic metrics, BLEU~\cite{papineni2002} and METEOR~\cite{BanerjeeLavie2005}. 
Table~\ref{tab:evalMT} shows the scores in two cases: \textit{all}, when the complete texts are evaluated and \textit{coref}, when only the subset of sentences that have been augmented in S3 are considered -- 265 out of 494 for news and 239 out of 518 for TED. For news, the best system is that trained on more data, S2; but for TED talks S3 with less data has the best performance.

The difference between the behaviour of the systems can be related to the different genres. We have seen that news are dominated by nominal mentions while TED is dominated by pronominal ones.  
Pronouns mostly need coreference information to be properly translated, while noun phrases can be improved simply because more instances of the nouns appear in the training data. With this, S3 improves the baseline S1 in +1.1 BLEU points for  {TED}$_{coref}$ but -0.2 BLEU points for news$_{coref}$.

However, even if the systems differ in the overall performance, the change is not related to the number of errors in coreference chains. Table~\ref{tab:evalMT} also reports the number of mistakes in the translation of coreferent mentions. Whereas the number of errors correlates with translation quality (as measured by BLEU) for news$_{coref}$ this is not the case of TED$_{coref}$.

\subsection{Error analysis}\label{sec:error}

\begin{figure*}[t]
\centering
\includegraphics[width=1.89\columnwidth]{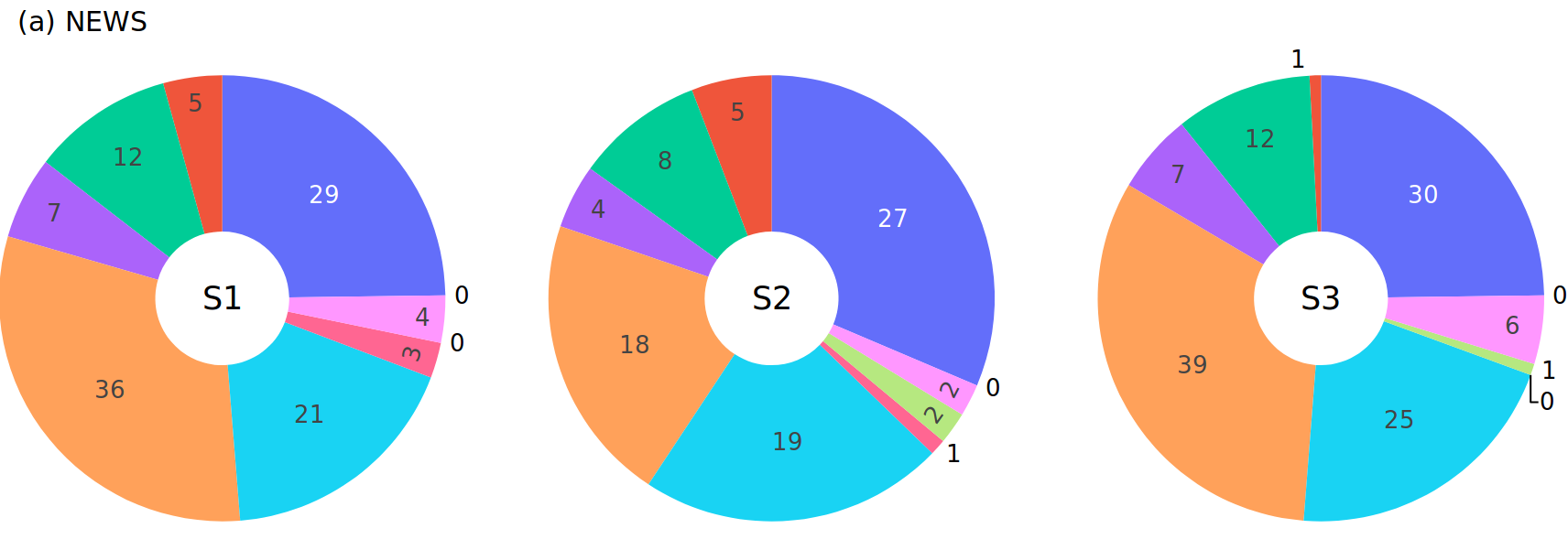}\\
\includegraphics[width=1.89\columnwidth]{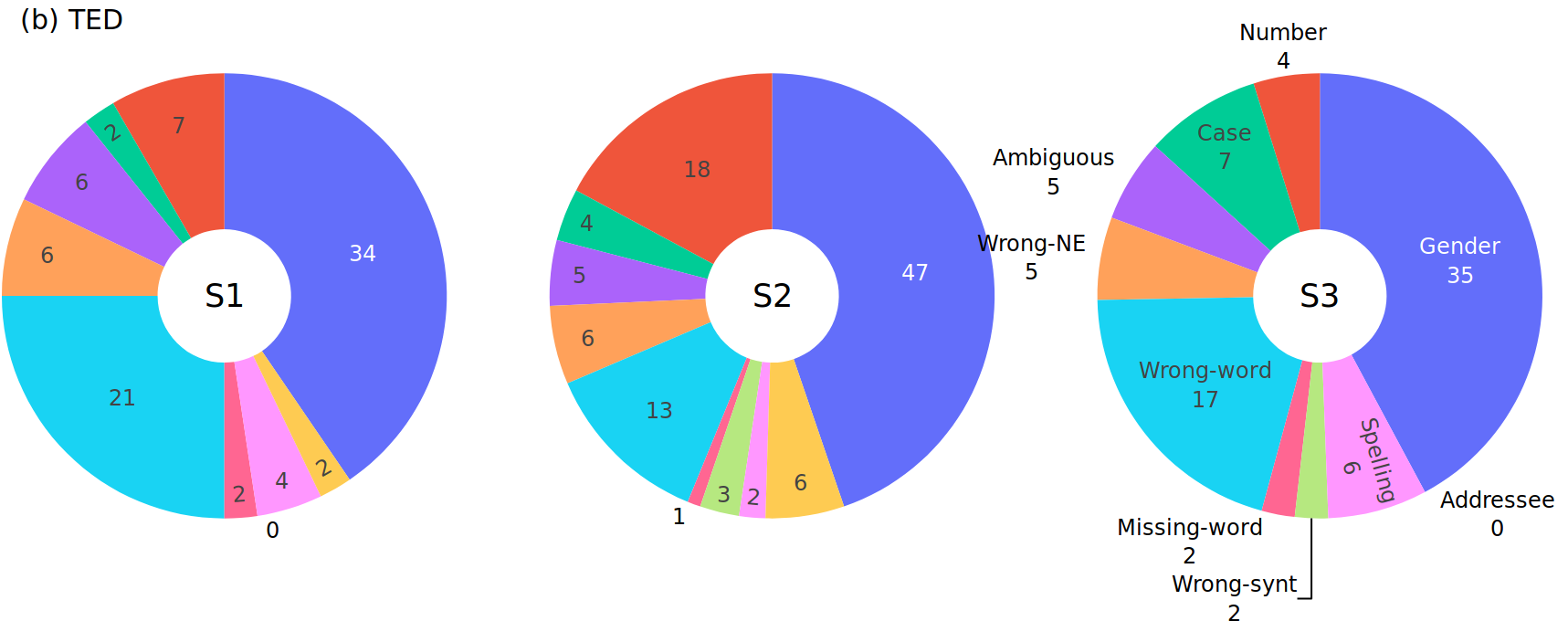}
\vspace*{-0.7em}
\caption{Number of errors per system (S1, S2, S3) and genre (news, TED). Notice that the total number of errors differs for each plot, total numbers are reported in Table~\ref{tab:evalMT}. Labels in Figure (b)--S3 apply to all the chart pies that use the same order and color scale for the different error types defined in Section~\ref{sec:error}.}
\label{fig:errors}
\end{figure*}

The total distribution for the 10 categories of errors defined in Section~\ref{sec:anno} can be seen in Figure~\ref{fig:errors}. Globally, the proportion of errors due to our closed categories (gender, number, case and ambiguous) is larger for TED talks than for news (see analysis in Section~\ref{sec:freqerror}).
Gender is an issue with all systems and genres which does not get solved by the addition of more data.
Additionally, news struggle with wrong words and named entities; for this genre the additional error types (see analysis in Section~\ref{sec:additionalerrors}) represent around 60\% of the errors of S1/S3 to be compared to the 40\% of TED talks. 

\subsubsection{Predefined error categories}\label{sec:freqerror}
 \setlength{\Exlabelsep}{0.4em}%
 \setlength{\Extopsep}{0.4\baselineskip}%
Within our predefined closed categories (gender, number, case and ambiguous), the gender errors belong to the most frequent errors. They include wrong gender translation of both pronouns, as {\sl sie} (``her'') instead of {\sl ihn} (``him'') in example~\ref{ex:gender-intersent} referring to the masculine noun {\sl Mindestlohn}, and nominal phrases, as {\sl der Stasi} instead of {\sl die Stasi}, where a masculine form of the definite article is used instead of a feminine one, in example~\ref{ex:gender-intrasent}.

\ex.\label{ex:gender-intersent}
src: {\sl [The current minimum wage] of 7.25 US dollars is a pittance... 
She wants to raise [it] to 15 dollars an hour.}\\
S3: {\sl [Der aktuelle Mindestlohn] von 7,25 US-Dollar sei Almosen... 
Sie m\"ochte [sie] auf 15 Dollar pro Stunde erhöhen.}

\ex.\label{ex:gender-intrasent}
src: {\sl 
...let's have a short look at the history of [the Stasi], because it is really important for understanding [its] self-conception.}\\
S2: {\sl Lassen sie uns... 
einen kurzen Blick auf die Geschichte [des Stasi] werfen denn es wirklich wichtig, [seine] Selbstauffassung zu verstehen.}


The gender-related errors are common to all the automatic translations. 
Interestingly, systems S1 and S3 have more problems with gender in translations of TED talks, whereas they do better in translating news, which leads us to assume that this is a data-dependent issue: while the antecedent for news is in the same sentence it is not for TED talks.
A closer look at the texts with a high number of gender problems confirms this assumption ---they contain references to females who were translated with male forms of nouns and pronouns (e.g. {\sl Mannschaftskapit\"an} instead of {\sl Mannschaftskapit\"anin}).  


We also observe errors related to gender for the cases of explicitation in translation. Some impersonal English constructions not having direct equivalents in German 
are translated with personal constructions, which requires an addition of a pronoun. Such cases of explicitation were automatically detected in parallel data in~\citep{LapshinovaHardmeier2017,LapshinovaEtAl2019}.
They belong to the category of obligatory explicitation, i.e. explicitation dictated by differences in the syntactic and
semantic structure of languages, as defined by~\citet{Klaudy2008}. An MT system tends to insert a male form instead of a female one even if it's marked as feminine (S3 adds the feminine form \emph{she} as markup), as illustrated in example~\ref{ex:gender-explicitation} where the automatic translation contains the masculine pronoun {\sl er} (``he'') instead of {\sl sie} (``she'').

\ex. \label{ex:gender-explicitation}
src: {\sl [Biles] earned the first one on Tuesday while serving as the exclamation point to retiring national team coordinator Martha Karolyi's going away party.}\\
ref: {\sl [Biles] holte die erste Medaille am Dienstag, w\"ahrend [sie] auf der Abschiedsfeier der sich in Ruhestand begehenden Mannschaftskoordinatorin Martha Karolyi als Ausrufezeichen diente.}\\
S2: {\sl [Biles] verdiente den ersten am Dienstag, w\"ahrend [er] als Ausrufezeichen f\"ur den pensionierten Koordinator der Nationalmannschaft, Martha Karolyi, diente.}

Another interesting case of a problem related to gender is the dependence of the referring expressions on grammatical restrictions in German. In example~\ref{ex:gender-gram}, the source chain contains the pronoun {\sl him} referring to both {\sl a 6-year-old boy} and {\sl The child}. In German, these two nominal phrases have different gender (masculine vs. neutral). The pronoun has grammatical agreement with the second noun of the chain ({\sl des Kindes}) and not its head ({\sl ein 6 Jahre alter Junge}).

\ex.\label{ex:gender-gram}
src: {\sl Police say [a 6-year-old boy] has been shot in Philadelphia... [The child]'s grandparents identified [him] to CBS Philadelphia as [Mahaj Brown].}\\
S1: {\sl Die Polizei behauptet, [ein 6 Jahre alter Junge] sei in Philadelphia erschossen worden... Die Gro{\ss}eltern [des Kindes] identifizierten [ihn] mit CBS Philadelphia als [Mahaj Brown].}

Case- and number-related errors are less frequent in our data. However, translations of TED talks with S2 contain much more number-related errors than other outputs. Example~\ref{ex:number} illustrates this error type which occurs within a sentence. The English source contains the nominal chain in singular {\sl the cost -- it}, whereas the German correspondence {\sl Kosten} has a plural form and requires a plural pronoun ({\sl sie}). However, the automatic translation contains the singular pronoun {\sl es}.

\ex. \label{ex:number}
src: {\sl ...to the point where [the cost] is now below 1,000 dollars, and it's confidently predicted that by the year 2015 [it] will be below 100 dollars...}\\
S2: {\sl bis zu dem Punkt, wo [die Kosten] jetzt unter 1.000 Dollar liegen, und es ist zuversichtlich, dass [es] bis zum Jahr 2015 unter 100 Dollar liegen wird...}



Ambiguous cases often contain a combination of errors or they are difficult to categorise due to the ambiguity of the source pronouns, as the pronoun {\sl it} in example~\ref{ex:ambiguous} which may refer either to the noun {\sl trouble} or even the clause {\sl Democracy is in trouble} is translated with the pronoun {\sl sie} (feminine). In case of the first meaning, the pronoun would be correct, but the form of the following verb should be in plural. In case of a singular form, we would need to use a demonstrative pronoun {\sl dies} (or possibly the personal pronoun {\sl es}).

\ex.\label{ex:ambiguous}
src: {\sl Democracy is in trouble... 
and [it] comes in part from a deep dilemma...}\\
S2: {\sl Die Demokratie steckt in Schwierigkeiten ... 
und [sie] r\"uhrt teilweise aus einem tiefen Dilemma her...}





\subsubsection{Additional error types}\label{sec:additionalerrors}

At first glance, the error types discussed in this section do not seem to be related to coreference ---a wrong translation of a noun can be 
traced back to the training data available and the way NMT deals with unknown words. However, a wrong translation of a noun may result in its invalidity to be a referring expression for a certain discourse item. As a consequence, a coreference chain is damaged. We illustrate a chain with a wrong named entity translation in example~\ref{ex:wrongne}. The source chain contains five nominal mentions referring to an American gymnast Aly Raisman: {\sl silver medalist} -- {``Final Five'' teammate} --  {\sl Aly Raisman} -- {\sl Aly Raisman} -- {\sl Raisman}. All the three systems used different names. Example~\ref{ex:wrongne} illustrates the translation with S2, where {\sl Aly Donovan} and {\sl Aly Encence} were used instead of {\sl Aly Raisman}, and the mention {\sl Raisman} disappears completely from the chain.

\ex.\label{ex:wrongne}
src: {\sl Her total of 62.198 was well clear of [silver medalist] and [``Final Five'' teammate] [Aly Raisman]...United States' Simone Biles, left, and [Aly Raisman] embrace after winning gold and silver respectively... [Raisman]'s performance was a bit of revenge from four years ago, when [she] tied...}\\
S2: {\sl Ihre Gesamtmenge von 62.198 war deutlich von [Silbermedaillengewinner] und [``Final Five'' Teamkollegen] [Aly Donovan]... Die Vereinigten Staaten Simone Biles, links und [Aly Encence] Umarmung nach dem Gewinn von Gold und Silber... Vor vier Jahren, als [sie]...}

Example~\ref{ex:wrongword1} illustrates translation of the chain {\sl The scaling in the opposite direction} -- {\sl that scale}. The noun phrases {\sl Die Verlagerung in die entgegengesetzte Richtung} (``the shift in the opposite direction'') and {\sl dieses Ausma{\ss}} (``extent/scale'') used in the S1 output do not corefer (cf. {\sl Wachstum in die entgegengesetzte Richtung} and {\sl Wachstum} in the reference translation). Notice that these cases with long noun phrases are not tackled by S3 either.

\ex.\label{ex:wrongword1}
src: {\sl [The scaling in the opposite direction]...
drive the structure of business towards the creation of new kinds of institutions that can achieve [that scale].}\\
ref: {\sl [Wachstum in die entgegengesetzte Richtung]... 
steuert die Struktur der Gesch\"afte  in Richtung Erschaffung von neuen Institutionen, die [dieses Wachstum] erreichen k\"onnen.}\\
S1: {\sl [Die Verlagerung in die entgegengesetzte Richtung]... 
treibt die Struktur der Unternehmen in Richtung der Schaffung neuer Arten von Institutionen, die [dieses Ausma{\ss}] erreichen k\"onnen.}

\subsubsection{Types of erroneous mentions}
Finally, we also analyse the types of the mentions marked as errors. They include either nominal phrases or pronouns.  Table~\ref{tab:errorcontext} shows that there is a variation between the news texts and TED talks in terms of these features. News contain more erroneous nominal phrases, whereas TED talks contain more pronoun-related errors. Whereas both the news and the TED talks have more errors in translating anaphors, there is a higher proportion of erroneous antecedents in the news than in the TED talks.

\begin{table}[t!]
    \centering
    \begin{tabular}{llc cc c cc}
    \hline
&	& &\bf ant.&\bf	ana.	& & \bf NP &\bf	pron.\\
	\hline
\bf news&\bf S1&& 0.30&	0.70&	&0.72&	0.28\\
\bf news&\bf S2&& 0.39&	0.61&	&0.63&	0.37\\
\bf news&\bf S3&& 0.36&	0.64&	&0.63&	0.37\\
\hline
\bf TED&\bf S1&& 0.18&	0.82&	&0.36&	0.64\\
\bf TED&\bf S2&& 0.18&	0.82&	&0.34&	0.66\\
\bf TED&\bf S3&& 0.28&	0.72&	&0.46&	0.54\\
\hline
    \end{tabular}
    \caption{Percentage of erroneous mentions: antencedent vs. anaphor, and noun phrase vs. pronominal. }
    \label{tab:errorcontext}
\end{table}


It is also interesting to see that S3 reduces the percentage of errors in anaphors for TED, but has a similar performance to S2 on news. 

\section{Summary and Conclusions}
\label{sec:conclusions}

We analysed coreferences in the translation outputs of three transformer systems that differ in the training data and in whether they have access to explicit intra- and cross-sentential anaphoric information (S3) or not (S1, S2). We see that the translation errors are more dependent on the genre than on the nature of the specific NMT system: whereas news  (with mainly NP mentions) contain a majority of errors related to wrong word selection, TED talks (with mainly pronominal mentions) are prone to accumulate errors on gender and number.


System S3 was specifically designed to solve this issue, but we cannot trace the improvement from S1 to S3 by just counting the errors and error types, as some errors disappear and others emerge: coreference quality and automatic translation quality do not correlate in our analysis on TED talks. As a further improvement to address the issue, we could add more parallel data to our training corpus with a higher density of coreference chains such as movie subtitles or parallel TED talks.

We also characterised the originals and translations according to coreference features such as total number of chains and mentions, average chain length and size of the longest chain. We see how NMT translations increase the number of mentions about $30\%$ with respect to human references showing even a more marked explicitation effect than human translations do.
As future work, we consider a more detailed comparison of the human and machine translations, and analyse the purpose of the additional mentions added by the NMT systems. It would be also interesting to evaluate of the quality of the automatically computed coreferences chains used for S3. 

\section*{Acknowledgments}
The annotation work was performed at Saarland University. We thank Anna Felsing, Francesco Fernicola, Viktoria Henn, Johanna Irsch, Kira Janine Jebing, Alicia Lauer, Friederike Lessau and Christina Pollkl\"asener for performing the manual annotation of the NMT outputs. The project on which this paper is based was partially funded by the German Federal Ministry of Education and Research under the funding code 01IW17001 (Deeplee) and by the German Research Foundation (DFG) as part of SFB 1102 Information Density and Linguistic Encoding. Responsibility for the content of this publication is with the authors.

\bibliography{corefNMT}
\bibliographystyle{acl_natbib}



\end{document}